\pdfoutput=1

\documentclass[11pt]{article}

\usepackage{acl2023}

\usepackage{times}
\usepackage{latexsym}

\usepackage[T1]{fontenc}

\usepackage[utf8]{inputenc}

\usepackage{microtype}
\usepackage{booktabs}
\usepackage{inconsolata}
\usepackage{amsmath}
\usepackage{amsfonts}
\usepackage{multirow}
\usepackage{array}
\usepackage{graphicx} 
\graphicspath{ {images/} }
\usepackage{caption}
\usepackage{subcaption}
\usepackage{hyperref}
\usepackage[compact]{titlesec}
\titlespacing{\section}{0pt}{2ex}{1ex}
\titlespacing{\subsection}{0pt}{1ex}{0ex}
\titlespacing{\subsubsection}{0pt}{0.5ex}{0ex}
\usepackage{paralist}

\usepackage[many]{tcolorbox}  
\newtcolorbox{boxH}{
    sharpish corners,
    boxrule = 0pt, 
    leftrule = 6pt 
}
\newtcolorbox{boxJ}{
    sharpish corners, 
    boxrule = 0pt, 
    toprule = 4.5pt, 
    enhanced,
    fuzzy shadow = {0pt}{-2pt}{-0.5pt}{0.5pt}{black!35} 
}

\title{States Hidden in Hidden States: \\Implicit Discrete State Representations Emerge in LLMs' Hidden States}

\author{Junhao Chen, Shengding Hu,  Zhiyuan Liu$^*$, Maosong Sun$^*$ \\
        Department of Computer Science and Technology, Tsinghua University \\ Beijing, China \\
        \texttt{chenjunh22@mails.tsinghua.edu.cn}}

\begin{document}
\maketitle
\def\thefootnote{$^*$}\footnotetext{Corresponding authors}
\begin{abstract}

Large Language Models (LLMs) exhibit emergent abilities that may reveal aspects of their internal mechanisms. We study one such capability: directly performing extended sequences of calculations without generating chain-of-thought solutions. The strongest models in our evaluation can directly output sums with up to 15 addends, where operands are sampled from 1 to 100. We hypothesize that models form Implicit Discrete State Representations (IDSRs) within their hidden states and use them for internal symbolic calculation. We test for these representations, characterize their formation from layer, digit, and sequence perspectives, and investigate their use in producing answers. We also find that these state representations are far from lossless in current open-source models, contributing to errors in final outputs. Our work offers an initial exploration of LLMs' symbolic calculation abilities and underlying mechanisms. Code and reproducibility artifacts are available at \url{https://github.com/Junhaoo-Chen/IDSR}.



\end{abstract}

\section{Introduction}

LLMs have demonstrated remarkable performance in a variety of fields \citep{Achiam2023GPT4TR,Touvron2023LLaMAOA}, including natural language understanding and generation \cite{Zhao2023ASO}, code generation \citep{Chen2021EvaluatingLL, Nijkamp2022CodeGenAO, Li2023StarCoderMT}, and mathematical problem-solving \cite{Hendrycks2021MeasuringMP}. These abilities emerge as the model scales. 


In this study, we dive into another intriguing emergent capability: the ability of LLMs to perform arithmetic calculations, particularly iterative additions directly, without relying on chain-of-thought reasoning. For example, given the question: "\textit{Please directly give me the answer to 17 + 38 + 32 + 87 + 47 + 28 + 17 + 21 + 53 + 15 + 18 + 76}", a SOTA LLM can directly produce the correct answer "449" without producing any intermediate tokens. This phenomenon warrants investigation for two principal reasons. Firstly, it is unlikely that models were trained on such iterative addition data, as it exerts negligible influence on overall performance across general domains and benchmarks~\cite{Wang2021GeneralizingTU}. This phenomenon likely emerges naturally during the scaling process and presents a more meaningful study subject compared to tasks that may have more intricate relations with memorizing training data. Secondly, the simplicity of this phenomenon renders it an ideal candidate for interpretability research, potentially serving as a foundational step in uncovering the internal mechanisms underlying LLMs in performing intrinsic iterative reasoning.

Prior research on the interpretability of models performing mathematical tasks focuses primarily on binary arithmetic operations~\cite{Zhu2024LanguageMU}. However, this body of work fails to explain the formation of discrete state representations within the hidden layers of these models. 

In this paper, we propose a central hypothesis to elucidate the emergent capability of implicit sequential computation: LLMs inherently track discrete states. By forming Implicit Discrete State Representations (IDSRs) that encapsulate intermediate results, LLMs can leverage these precomputed intermediate results for subsequent use, thereby preventing the necessity for intricate computations in the final step.




To validate this hypothesis, we construct a synthetic dataset of iterative addition problems and employ probing methods to examine the existence of IDSRs in hidden states across various LLMs. Upon confirming its existence, we further investigate the properties and formation of IDSRs, and demonstrate its formation through digit-wise, layer-wise, and sequence-wise perspectives, and provide noteworthy observations of distinct layer functionalities. From a digit-wise perspective, IDSRs form independently and sequentially, beginning with the lowermost digit. From the layer-wise level, a sharp transition from shallow semantic computation to semantic understanding occurs around layer 10, and a shift from linearity to non-linearity arises in the later model layers. From a sequence-wise perspective, information encoded in IDSRs is propagated along the sequence for sequential utilization. Finally, we confirm that the model utilizes IDSRs to produce the final result rather than computing using all preceding numbers simultaneously. This investigation provides insight into the multi-step reasoning and state-tracking abilities of LLMs~\cite {Singh2024RethinkingII, Li2023InferenceTimeIE}.

\section{Related Work}

\paragraph{LLMs' State Tracking Abilities.}

Language models are exhibiting increasingly mature abilities to perform arithmetic tasks, both open-sourced \citep{Dubey2024TheL3, DeepSeekAI2024DeepSeekV3TR, Yang2024Qwen25TR} and close-sourced models \citep{Open_AI_2024, team2023gemini, Anthropic_2024} are excelling at a variety of mathematical benchmarks, ranging from elementary to Olympic difficulty levels \citep{Hendrycks2021MeasuringMP, Chen2023TheoremQAAT, Li2024GSMPlusAC, He2024OlympiadBenchAC}. 


Other abilities that are discussed a lot are LMs' state encoding and tracking abilities \citep{Merrill2024TheIO}. \citeauthor{Li2022EmergentWR} and \citeauthor{Nanda2023EmergentLR} investigate the existence of non-linguistic state representations in board game settings, while \citeauthor{li2021implicit} find that model representations also encode entity states in the process of textual tasks. Taking this problem further, \citeauthor{Kim2023EntityTI} and \citeauthor{Dziri2023FaithAF} show that models perform non-trivial state tracking given specific composite tasks. However, whether LMs track discrete states during arithmetic tasks still remains an open question.

\paragraph{Interpretability of LLMs' Arithmetic Abilities.}

The inner workings of LMs in performing arithmetic and reasoning tasks are under-explored. Current literature suggests that neurons and layers inside LLMs may serve as feature extractors, extracting latent properties from inputs and passing them through layers \citep{Mikolov2013LinguisticRI, belinkov2022probing, Burns2022DiscoveringLK, Gurnee2023FindingNI}. 

Building on this idea, recent work demonstrates that hidden states during inference contain representations relevant to future tokens \citep{Nostalgebraist_2020, Belrose2023ElicitingLP, Pal2023FutureLA, wu2024language}. This insight underpins our research, in which we prove the existence and utilization of implicit representations in LMs.

Previous analyses have also examined the arithmetic capabilities of LMs. \citeauthor{Stolfo2023AMI}, \citeauthor{Qiu2024DissectingMI} and \citeauthor{Zhang2024InterpretingAI} identify that LMs employ MLPs and attention heads at different stages of arithmetic reasoning, while \citeauthor{Kantamneni2025LanguageMU} provides explanations based on trigonometry.

\paragraph{Broader Interpretability of LLMs.} Multiple paradigm has emerged to investigate the working mechanisms of LLMs. Among them, the most famous ones are mechanistic interpretability~\cite{NEURIPS2023_34e1dbe9, elhage2021mathematical} and representation engineering~\cite{Zou2023RepresentationEA}. 

Mechanistic interpretability proposes to check the neuron-level activations and understand the functioning circuits of LLMs. From this perspective, mathematical tasks have been widely used as a tool because of their simplicity. For example, it is used to discover grokking in~\citet{varma2023explaining} and understand double descent and emergent abilities in~\citet{huang2024unified}. However, these works do not focus on the mathematical ability that emerges in SOTA LLMs.


Representation engineering is another pivotal approach in model interpretability, emphasizing the holistic feature representations within layers \citep{li2021implicit, Zou2023RepresentationEA}. This approach facilitates behavioral monitoring and performance modification \citep{Zhang2024TowardsGC, Li2023InferenceTimeIE}. However, it is still underdeveloped in practical applications. A simple form of representation engineering is the widely used approach "probing" \citep{alain2016understanding, belinkov2022probing, Hernandez2023LinearityOR}, which typically involves using simple auxiliary models to make classifications. 
Research in this field extends to specific scenarios. \citeauthor{Li2022EmergentWR} and \citeauthor{Nanda2023EmergentLR} examine board game, yielding divergent conclusions regarding the linearity of hidden states. \citeauthor{Yang2024DoLL} explores event reasoning, demonstrating that reasoning predominantly occurs in the initial inference step and scales with model size. 

Few studies, however, critically assess the arithmetic capabilities of LMs. Some examine neuron activations only \citep{Stolfo2023AMI}, while others focus on simple calculations without comprehensively considering model layers \citep{Zhu2024LanguageMU}. This paper also studies the LLMs' interpretability with probing techniques but focuses on the unique perspective of IDSR. 


\begin{figure}[t]
    \centering
    \includegraphics[width=0.88\linewidth]{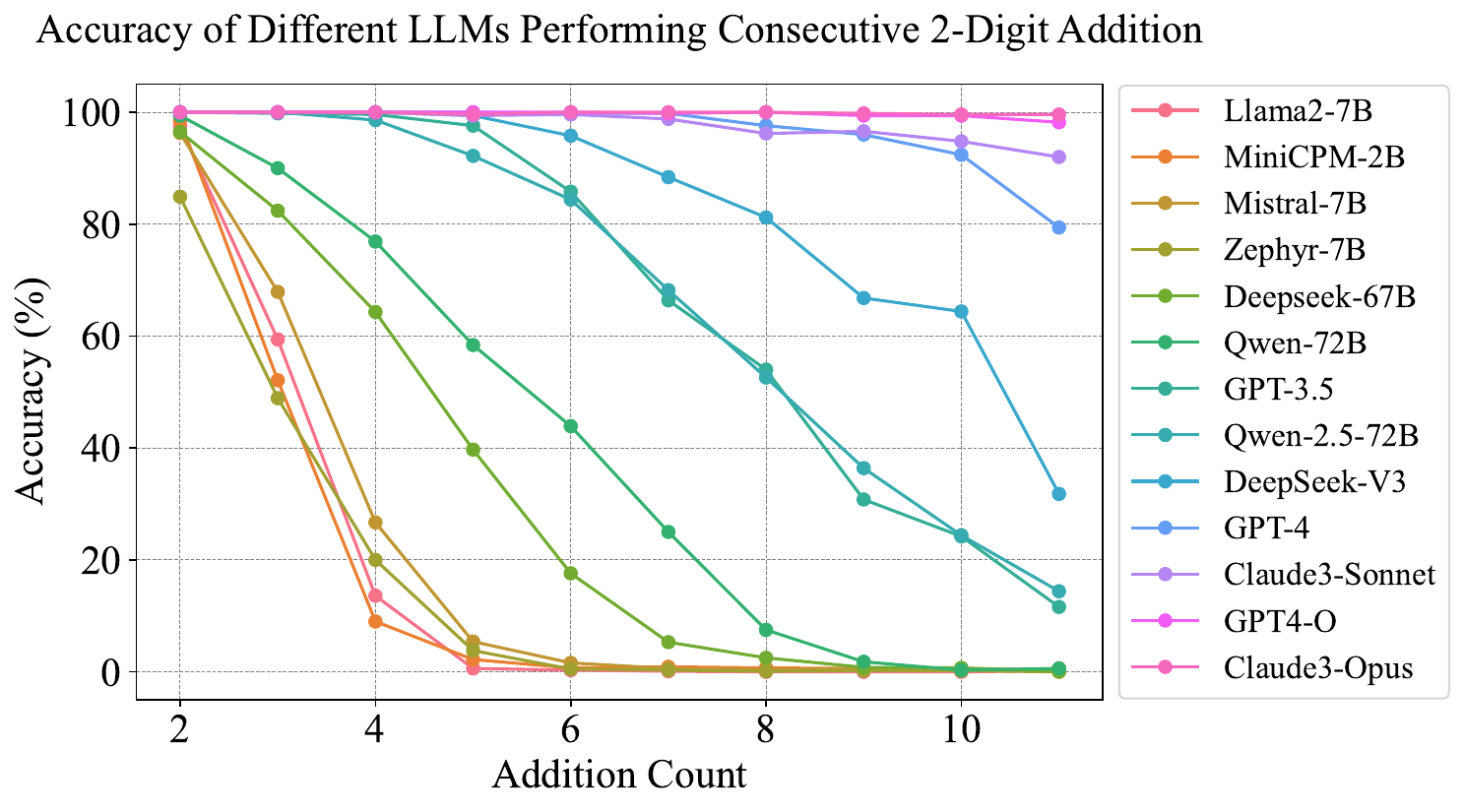}
    \caption{Accuracies of Different Models Performing Iterative Addition with Operands Sampled from 1 to 100}
    \label{fig:models}
    \vspace{-9pt}
\end{figure}

\begin{figure}[t]
    \centering
    \includegraphics[width=0.82\linewidth]{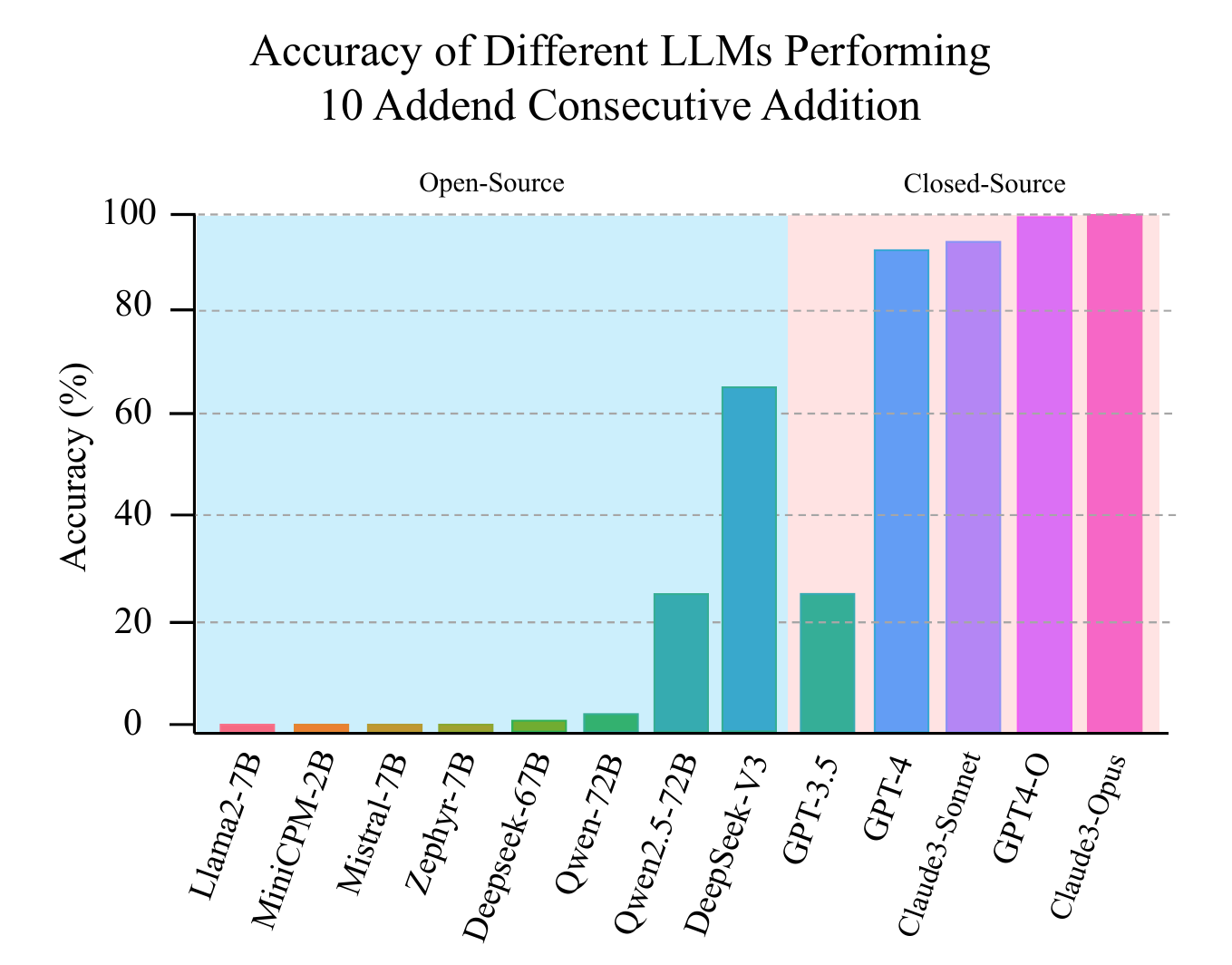}
    \caption{Accuracies of Different Models Performing 10-Addend Addition with Operands Sampled from 1 to 100}
    \label{fig:emergent}
    \vspace{-12pt}
\end{figure}

\section{Emergence of Implicit Computation}
\label{sec:ability}

In this section, we confirm the emergent abilities of \emph{implicit computation} using a variety of both open-source and closed-source LLMs.

\paragraph{Problem Statement.}
We employ implicit iterative addition as the representative task. In this context, the model is tasked with delivering the sum of an extended sequence of additions directly. An example prompt is provided below:

\begin{boxJ}
\textit{Please directly give me the answer to 17 + 38 + 32 + 87 + 47 + 28 + 17 + 21 + 53 + 15 + 18 + 76.}
\end{boxJ}

There are three reasons why the ability to solve such a task might indicate the formation of discrete state representations:

\begin{compactenum}
    \item This capability is unlikely to be a result of memorizing existing training data, as storing the results of calculations necessitates a parameter space of $O(99^L)$.
    
    \item Direct optimization of this task during training is unlikely. As Goodhart's law~\cite{Strathern1997} suggests, "When a measure becomes a target, it ceases to be a good measure." Iterative addition offers minimal practical performance benefits, rendering it an unlikely optimization target. As a result, this ability could indeed stem from extensive unsupervised training.
    
    \item Each computational step is relatively simple. We exclude addition involving four-digit or more due to its increased single-step complexity, which complicates tracing implicit computation because of single-step errors.
\end{compactenum}

To ensure that models that are only accessible through API calls do not rely on tools such as calculators, we manually verify that there is at least one addition count where the model has less than a 100\% probability of yielding the correct answer. Additionally, we ensure that the models do not utilize explicit chain-of-thought reasoning through prompt engineering.

Specifically, we evaluate the exact accuracy of the predicted answers against the ground truth for different models directly performing iterative addition of varying lengths from 2 to 14. For every model and addition count, operands are sampled from the integers 1 through 100, inclusive. We collect 1,000 accepted responses per condition under the same procedure: unparsable responses and responses whose relative deviation from the ground truth exceeds 200\% are resampled.

We include the following LLMs in our analysis: Llama2-7B~\cite{Touvron2023Llama2O}, MiniCPM-2B~\cite{Hu2024MiniCPMUT}, Mistral-7B~\cite{Jiang2023Mistral7}, Zephyr-7B~\cite{Tunstall2023ZephyrDD}, DeepSeek-67B~\cite{Bi2024DeepSeekLS}, Qwen-72B~\cite{Bai2023QwenTR}, the Qwen2.5 series with different sizes~\cite{Yang2024Qwen25TR}, and DeepSeek-V3~\cite{DeepSeekAI2024DeepSeekV3TR}. For closed-source LLMs, we consider GPT-3.5~\cite{Brown2020LanguageMA}, GPT-4~\cite{Achiam2023GPT4TR}, Claude3-Sonnet, Claude3-Opus~\cite{Anthropic_2024}, and GPT4-O~\cite{Open_AI_2024}. 

As illustrated in Figure \ref{fig:models}, there exists a strong correlation between performance and model size. Smaller models achieve passable accuracy when adding 2 or 3 operands, but their accuracy rapidly deteriorates to near zero when the length of the sequence reaches 8. Larger models, however, maintain accuracies above 50\% for sequences of up to 8 numbers and demonstrate non-zero performance for sequences of 11 or even more numbers, demonstrating the fast "emergence" of this capability.

The "emergence" of this capability becomes prominent when models encounter more than 8 addends. Figure \ref{fig:emergent} presents the accuracies of models performing direct addition with 10 addends. It is evident that larger and more advanced closed-source models exhibit higher accuracies emergently.

\begin{figure}
    \centering
    \includegraphics[width=\linewidth]{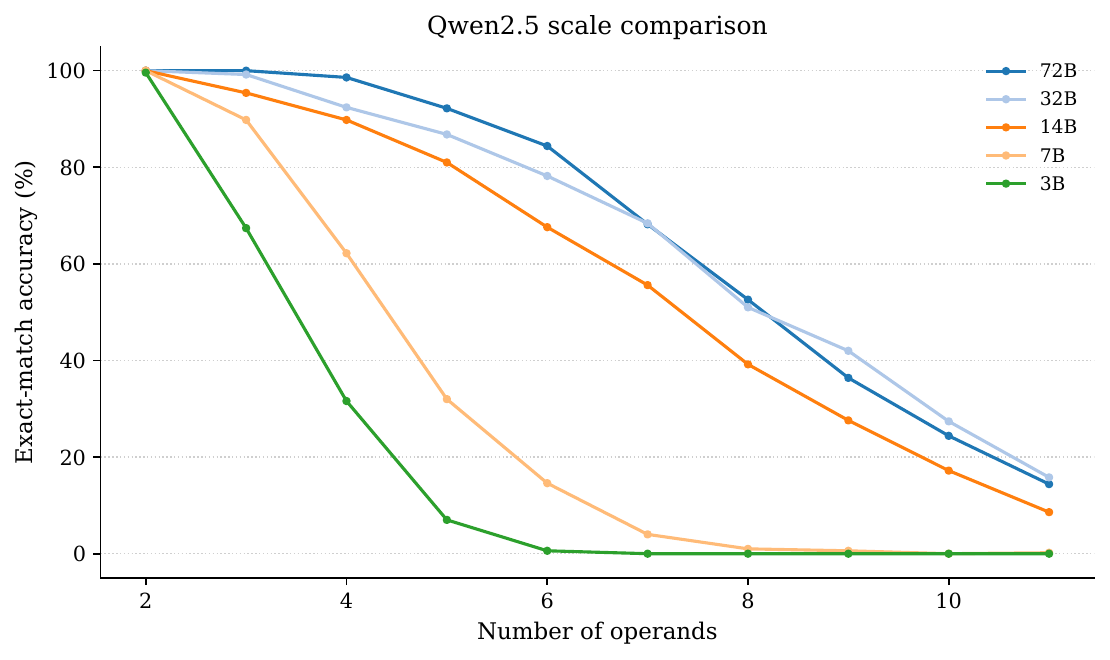}
    \vspace{-25pt}
    \caption{Accuracies of the Qwen2.5 Series Performing Iterative Addition with Operands Sampled from 1 to 100}
    \label{fig:qwen}
    \vspace{-20pt}
\end{figure}

To analyze the correlation between model size and performance, we examine the Qwen2.5 series, including models with sizes of 72B, 32B, 14B, 7B and 3B, as illustrated in Figure \ref{fig:qwen}. The results indicate a distinct enhancement in performance proportional to the increase in model size, especially noticeable for sequence lengths around five.



\section{Analysis Methodology}
Given the ability of models to directly yield calculation outcomes, we hypothesize that these models form \textbf{Implicit Discrete State Representations (IDSRs)} of intermediate results. For example, consider the formula $13 + 24 + 41 =$. We propose that the most plausible mechanism for models to complete this calculation in a single pass is to generate an IDSR of 37 ($13 + 24$) at the second "+" token, which would subsequently be utilized for the next step of computation (i.e., $+ 41$).

To thoroughly analyze this hypothesis, we propose and investigate these research questions:

\begin{boxH}
\label{box:RQ}
\begin{compactenum}
    \item[RQ1:] \textit{Do IDSRs really exist?}
    \item[RQ2:] \textit{What are IDSRs' properties?}
    \item[RQ3:] \textit{How do the IDSRs' form?}
    \item[RQ4:] \textit{How do models utilize IDSRs?}
\end{compactenum}
\end{boxH}

\subsection{Experiment Setup}

\subsubsection{Dataset }
\label{subsubsec:dataset}
We construct a straight-forward dataset of iterative addition and subtraction problems with different length, addend digits and prompts.

Our question prompts are divided into three categories, respectively formatted as in Table \ref{tab:math_expressions}, where \textit{i} ranges from 2 to 14, and $\{x_{i}\}$ are positive integers with the same number of digits ranging from 1 to 3. Prompts are chosen with a diversity of semantics to demonstrate the influence of context on IDSRs tracking.


\begin{table}[h]
    \centering
    \begin{tabular}{lp{4.8cm}}
         \toprule
         \textbf{Type} & \textbf{Expression} \\
         \midrule
         Addition & $\{x_{0}\}+\{x_{1}\}+...+\{x_{i-1}\}=$ \\
         \midrule
         Subtraction & $\{x_{0}\}+...+\{x_{i-2}\}-\{x_{i-1}\}=$ \\
         \midrule
         Prompting & \{Prompt\}, $\{x_{0}\}+\{x_{1}\}+...+\{x_{i-1}\}=$ \\
         \bottomrule
    \end{tabular}
    \caption{Dataset Expressions}
    \label{tab:math_expressions}
    \vspace{-12pt}
\end{table}

The dataset consists of 131,300 questions, as shown in Table \ref{tab:dataset_distribution}. Questions are sampled so that sums follow a uniform distribution, thereby eliminating probability bias and facilitating unbiased probe learning. The dataset is partitioned into training, validation, and test sets following an 80/10/10\% split for probing, respectively. 



\begin{table}[h]
    \centering
    \begin{tabular}{lcl}
         \toprule
         \textbf{Type} & \textbf{\#Digits} & \textbf{\#Questions} \\
         \midrule
         \multirow{3}{*}{Addition} & 3 & 39,000 \\
         & 2 & 6,500 \\
         & 1 & 1,300 \\
         \midrule
         \multirow{2}{*}{Subtraction}& 3 & 39,000 \\
         & 2 & 6,500 \\
         \midrule
         Prompting & 3 & 39,000 \\
         \bottomrule
    \end{tabular}
    \caption{Dataset Distribution}
    \label{tab:dataset_distribution}
    \vspace{-12pt}
\end{table}

\subsubsection{Hidden States }
\label{subsubsec:hiddenstates}
We prompt the model to answer dataset questions directly. During inference, we retrieve the hidden state $\textbf{H}_{i,j}$ corresponding to the $j^\text{th}$ token of the input sequence from layer \textit{i} of the model.

In our experiments, we exclusively extract the hidden states corresponding to the operator tokens for probing. This ensures that IDSRs are prominent and unbiased, as extracting IDSRs from operand tokens would incorporate representations of the operands themselves, introducing non-uniform bias and compromising the probing process.

\subsubsection{Classification Probes }
\label{subsubsec:probes}
Previous work has proven the abilities of probes on a wide variety of classification tasks. In our work, we utilize a multi-layer perceptron with one hidden layer to perform classification.

Specifically, the probing network is as follows:
\begin{equation}
\mathbf{P}_{i,j}^d = \operatorname{Softmax}\!\left(\mathbf{W}_{2}\,\sigma(\mathbf{W}_{1}\mathbf{H}_{i,j})\right)
\end{equation}
where $\mathbf{P}_{i,j}^d$ is the probing prediction for the $d^\text{th}$ digit of the IDSR, $\mathbf{W}_{1}\in\mathbb{R}^{d_{h}\times d_{m}}$ and $\mathbf{W}_{2}\in\mathbb{R}^{d_{o}\times d_{h}}$ are the probe weights, and $d_{m}$ and $d_{h}$ are the dimensions of the model and probe hidden states, respectively. We set $d_o=10$ because each probe predicts a digit from 0 to 9, and set $d_h=\sqrt{d_m d_o}$. 



In our experimental setup, three distinct types of probes are utilized: two multi-layer perceptrons with different hidden layer sizes and a single-layer perceptron. The respective parameter counts for each model type are detailed in Table \ref{tab:probe_sizes}.

For each experimental setting, probing models are trained on eight 80G A100 GPUs for a period ranging from 240 to 720 epochs. The duration depends on the specific input and the number of epochs required for convergence.

The learning rate is set to $1 \times 10^{-3}$, employing a stochastic gradient descent (SGD) optimizer. The model is optimized based on cross-entropy loss.

\begin{table}[htbp]
    \vspace{-10pt}
    \centering
    \begin{tabular}{cc}
         \toprule
         \textbf{Perceptron Model} & \textbf{Number of Parameters} \\
         \midrule
         Multi-Layer & 829,400 \\
         \midrule
         Multi-Layer \\ (Bottle-Necked)  & 81,920 \\
         \midrule
         Single-Layer & 40,960 \\
         \bottomrule
    \end{tabular}
    \caption{Probe Model Sizes}
    \label{tab:probe_sizes}
    \vspace{-10pt}
\end{table}


\subsubsection{Metrics }
\label{subsubsec:metrics}
For the assessment of model capabilities in performing iterative addition, we employ exact accuracy as our primary metric (\textbf{EA}, the ratio of the exact matches between the model output and the ground truth to the total number of questions).

To evaluate the classification probes, we compute the exact accuracy for each individual digit (\textbf{IDA}) as well as the overall exact accuracy (\textbf{OEA}, which considers a match only when all digits are predicted correctly).

\subsubsection{Models Chosen}
\label{subsubsec:models}
For our experimental setup, we select DeepSeek-67B \cite{Bi2024DeepSeekLS}, DeepSeek-V3 \cite{DeepSeekAI2024DeepSeekV3TR} and Qwen series models (4B, 7B, 14B, and 72B) \cite{Bai2023QwenTR} as representatives of open-source models. We aim to evaluate the proficiency in executing iterative addition tasks across a diverse range of models varying in size and capabilities. Special emphasis is placed on the Qwen-72B model to conduct an in-depth analysis of representation engineering and IDSRs' properties.

\section{Existence and Properties of IDSRs}
In this section, we present evidence of IDSRs regarding \hyperref[box:RQ]{RQ1} and \hyperref[box:RQ]{RQ2}. To demonstrate the existence of IDSRs in hidden states during inference, we design a series of probing prediction experiments with two levels of difficulty: Whole Number Probing and Digit-wise Probing.

\subsection{Whole Numbers Probing}
\label{subsubsec:whole_num_exist}

\begin{figure}
    \centering
    \begin{subfigure}[b]{1.0\linewidth}
        \centering
        \includegraphics[width=1.0\linewidth]{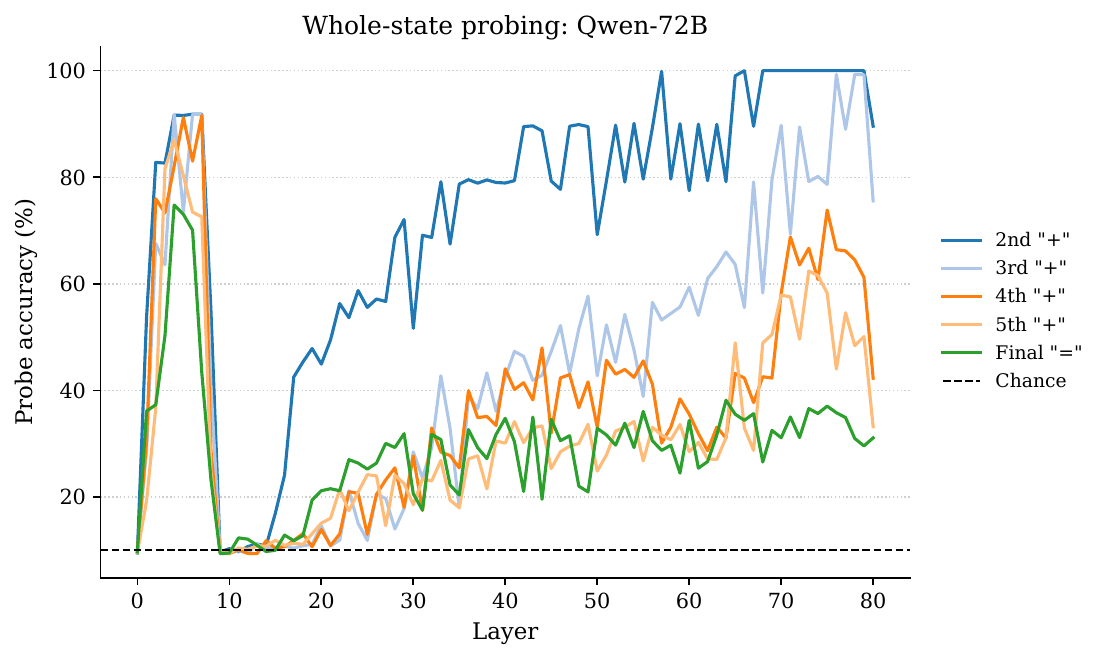}
        \vspace{-16pt}
        \caption{Qwen-72B}
        \vspace{-12pt}
        \label{fig:whole_pred_qwen}
    \end{subfigure}%
    \vspace{4mm}
    \begin{subfigure}[b]{1.0\linewidth}
        \centering
        \includegraphics[width=1.0\linewidth]{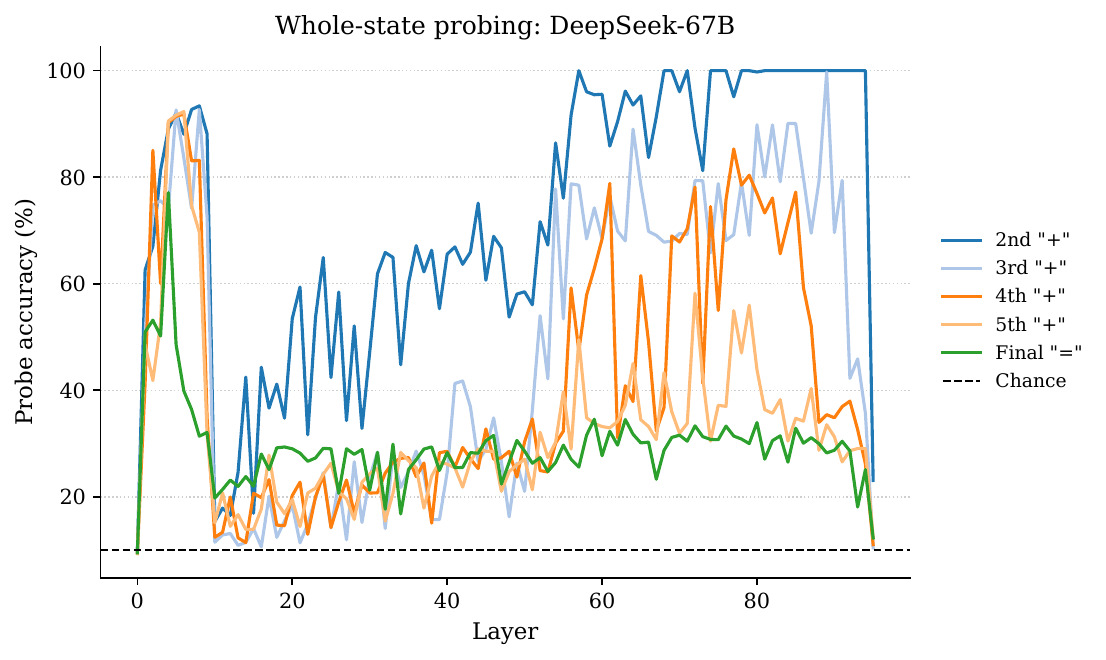}
        \vspace{-16pt}
        \caption{DeepSeek-67B}
        \vspace{-12pt}
        \label{fig:whole_pred_deepseek}
    \end{subfigure}
    \vspace{-8pt}
    \caption{Accuracies of Whole Number Probing Predictions}
    \label{fig:whole_pred}
    \vspace{-25pt}
\end{figure}

In this set of experiments, we train probes to predict the results as whole numbers from 10 possible sums. We probe different token positions across layers to investigate the existence of IDSRs' transference along the formula. The results, illustrated in Figure \ref{fig:whole_pred}, indicate that prediction accuracies significantly exceed random chance in all scenarios, demonstrating the existence of IDSR.

However, the process of forming IDSRs is far from lossless. The maximum prediction accuracies for the second to fifth addition signs and the final equal sign are 100\%, 99\%, 74\%, 62\%, and 37\% respectively, indicating substantial data and resolution loss as IDSRs are passed along the formula during inference. We hypothesize that reducing this error margin in the transference of IDSRs would enhance the capability of LLMs. This will be explored in future research.

Interesting trends across layers can also be observed in Figure~\ref{fig:whole_pred}, which will be discussed and analyzed in detail in Section~\ref{sec:formation}.

\subsection{Digit-wise Probing}
To investigate whether digits exist separately in IDSRs, we employ multiple probes to predict different digits of the sum. For this experiment, we choose formulas with 3-digit sums, necessitating the use of three probes. The range of possible sums for the $n_\text{th}$ addition/equal sign increases significantly, from 10 in the previous experiment setting to $\max\{999, 99n\} - \min\{100, 10n\}$, an increase of 10 to 40 times. We employ OEA as our primary metric, as mentioned in Section \ref{subsubsec:metrics}.

\begin{figure}
    \centering
    \includegraphics[width=0.9\linewidth]{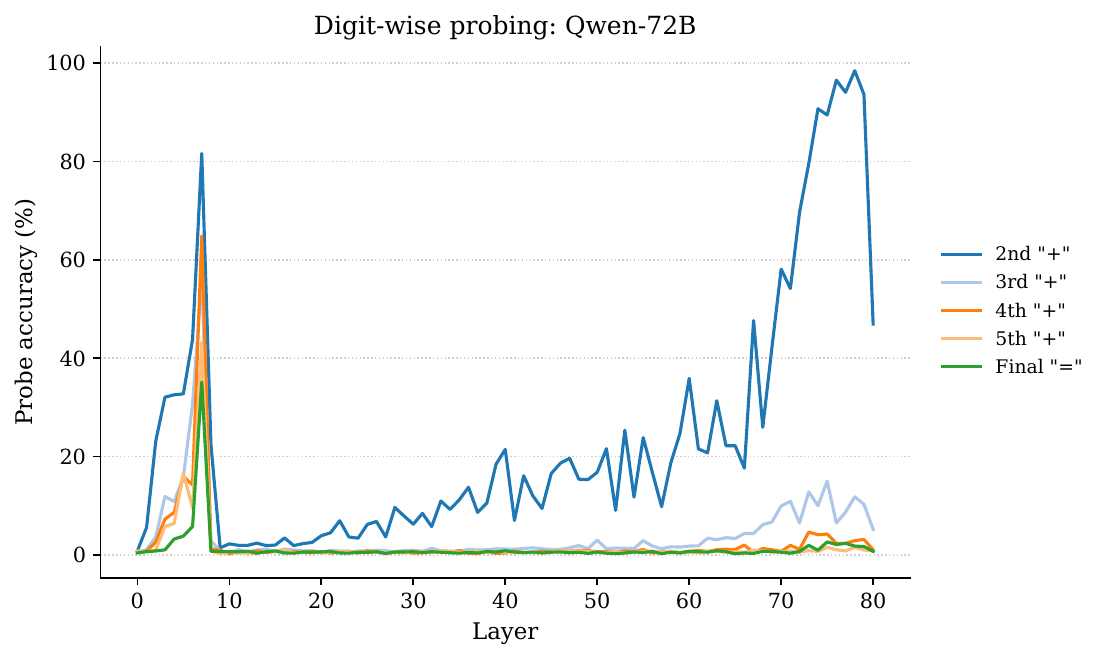}
    \vspace{-8pt}
    \caption{Accuracies of By-Digit Probing Predictions}
    \label{fig:qwen_2-digit_digit_acc}
    \vspace{-10pt}
\end{figure}

As depicted in Figure \ref{fig:qwen_2-digit_digit_acc}, probing accuracies using tokens from the first ten layers and the second addition sign from the later layers remain high. However, after significantly increasing prediction difficulty, the ability of probes to make exact predictions after the second addition declines sharply, indicating that models struggle to produce high-resolution IDSRs iteratively.

\subsection{Are IDSRs Linear?}
To gain a concrete understanding of IDSRs, we first examine its linearity. Beyond the original probing model with hidden size $\sqrt{d_{m}d_{o}}$, we construct 1) a smaller bottle-necked probing model with hidden size 10, as well as 2) a simpler single-layer perceptron utilizing a softmax activation function.
\begin{equation}
\mathbf{P}_{i,j}^d = \textit{Softmax}(\mathbf{W}_{1}\mathbf{H}_{i,j})
\end{equation}

\begin{figure}
    \centering
    \includegraphics[width=0.9\linewidth]{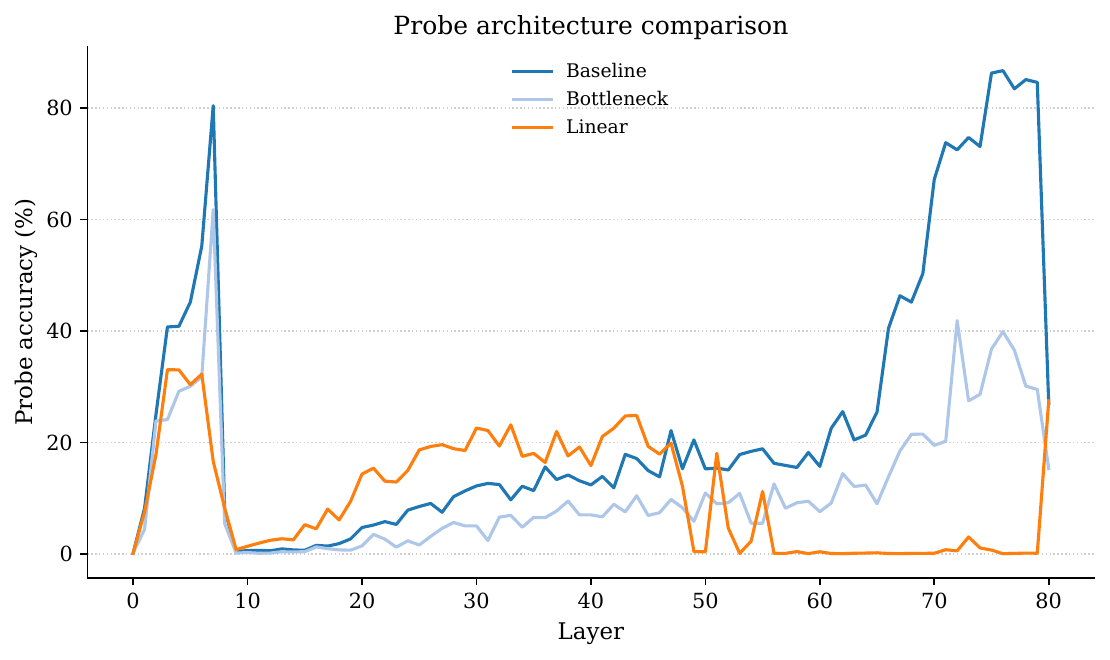}
    \vspace{-8pt}
    \caption{Accuracies of By-Digit Probing Predictions Using Different Probes}
    \label{fig:architecture_accuracies}
    \vspace{-12pt}
\end{figure}

As illustrated in Figure \ref{fig:architecture_accuracies}, layers 0 to 65 exhibit only minor accuracy drops with reduced hidden size, whereas layers 65 to 79 experience a significant reduction.

Notably, opposing accuracy trends appear in later layers for multi-layer and single-layer perceptrons. Between layers 50 and 65, accuracies for single-layer perceptrons drop to nearly zero, followed by a sharp increase for multi-layer perceptrons. This implies that layers 0 to 50 contain linear IDSRs, likely directions in the latent space. In contrast, layers 50 to 65 transit from linear to non-linear features, enhancing representation resolution and information density.

\section{Formation and Utilization of IDSRs}
\label{sec:formation}

In addition to analyzing the specific properties of IDSRs, we extend our study to overall formations. In this section, we identify patterns exhibited during inference at the digit level, sequence level, and layer level, revealing the inner mechanisms of iterative addition and multi-hop reasoning for LMs (\hyperref[box:RQ]{RQ3}). Following the formation analysis, we examine the utilization of such states (\hyperref[box:RQ]{RQ4}).

\subsection{Digit-Level Formation}

\begin{figure}
    \centering
    \begin{subfigure}[b]{1.0\linewidth}
        \centering
        \includegraphics[width=0.9\linewidth]{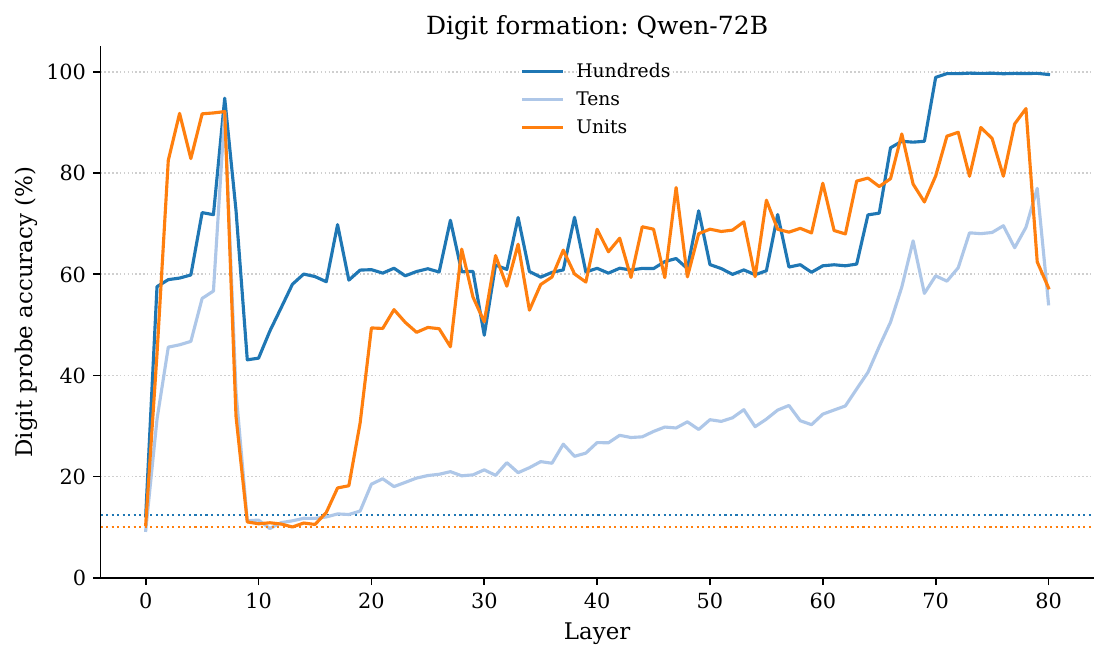}
        \vspace{-5pt}
        \caption{Qwen-72B}
        \vspace{-10pt}
        \label{fig:qwen_3-digit_digit_acc}
    \end{subfigure}%
    \vspace{4mm}
    \begin{subfigure}[b]{1.0\linewidth}
        \centering
        \includegraphics[width=0.9\linewidth]{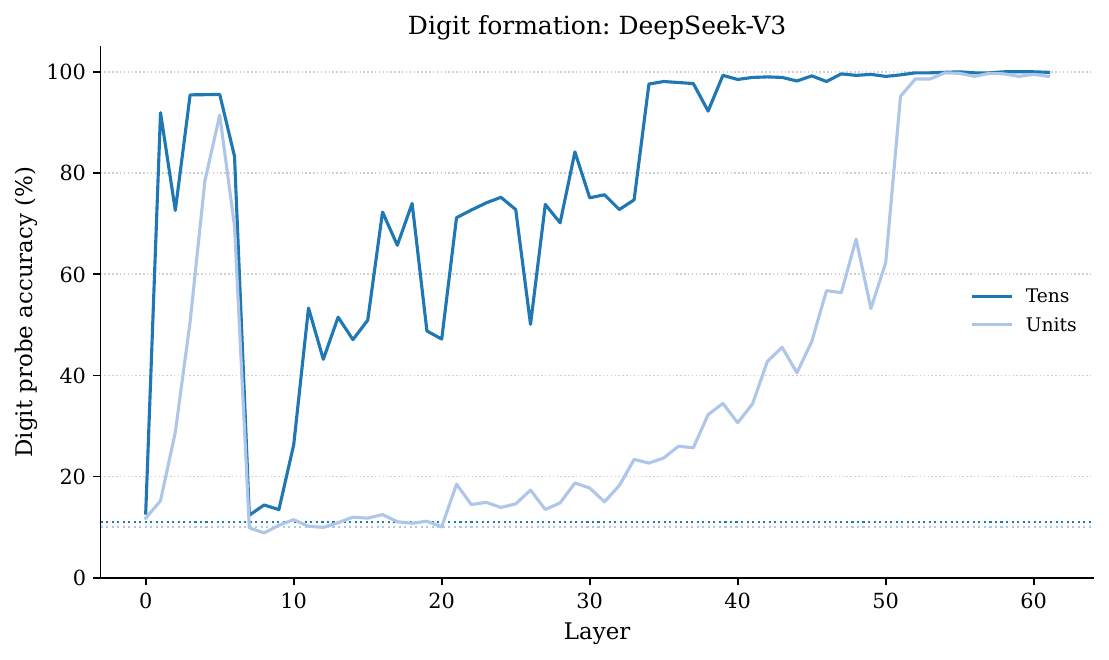}
        \vspace{-5pt}
        \caption{DeepSeek-V3}
        \vspace{-12pt}
        \label{fig:dpsk_v3_2_digit_digit_acc}
    \end{subfigure}
    \vspace{-8pt}
    \caption{Digit Accuracies of By-Digit Probing Predictions}
    \label{fig:digit_trends}
    \vspace{-15pt}
\end{figure}

We investigate the second addition operation within both three-digit and two-digit addition tasks, deriving two critical observations. First, the product of exact accuracies for the individual digits equals the overall exact accuracy, implying that models establish independent IDSRs. Second, as depicted in Figure \ref{fig:digit_trends}, the sequence in which digit prediction accuracies surpass random chance, as determined by statistical measures and annotated in the figure, follows an ascending digit order. This pattern mirrors the order humans use for digit-by-digit calculations, suggesting that models perform multi-digit addition through a series of iterative single-digit additions.


\subsection{Sequence-Level Formation}
\label{subsec: seqform}


The representation resolution of earlier addition sign tokens, as indicated by prediction accuracies, improves at earlier stages of the inference pass. The order of this resolution enhancement in Figures \ref{fig:whole_pred} and \ref{fig:qwen_2-digit_digit_acc} aligns precisely with the sequential order of the addition signs in the formula. This suggests that information encoded in IDSRs propagates along the sequence, allowing later tokens to utilize numerical IDSRs from earlier tokens for implicit calculations. In other words, \textbf{LLMs are performing iterative arithmetic tasks sequentially.}


\subsection{Layer-Level Formation}
\label{subsec:layer_form}




As depicted in Figures \ref{fig:whole_pred}, \ref{fig:qwen_2-digit_digit_acc}, and \ref{fig:architecture_accuracies}, an abrupt peak in IDSRs' resolution appears around the tenth layers for both models. Beyond this point, the resolution reinitializes from near non-existent levels.

We propose the hypothesis that the first ten layers employ a different mechanism from the later layers, particularly in multi-step reasoning tasks such as iterative addition. The first ten layers, termed "shallow-semantic layers", generate direct representations of arithmetic content regardless of the specific task. Conversely, the later layers, termed "semantic layers", incorporate task context, redoing the formation of the IDSRs in the process.

Utilizing the "subtraction" and "prompting" tasks discussed in Section~\ref{subsubsec:dataset}, we conduct two sets of experiments to demonstrate the existence of shallow-semantic and semantic layers.

\begin{figure}
    \centering
    \includegraphics[width=0.9\linewidth]{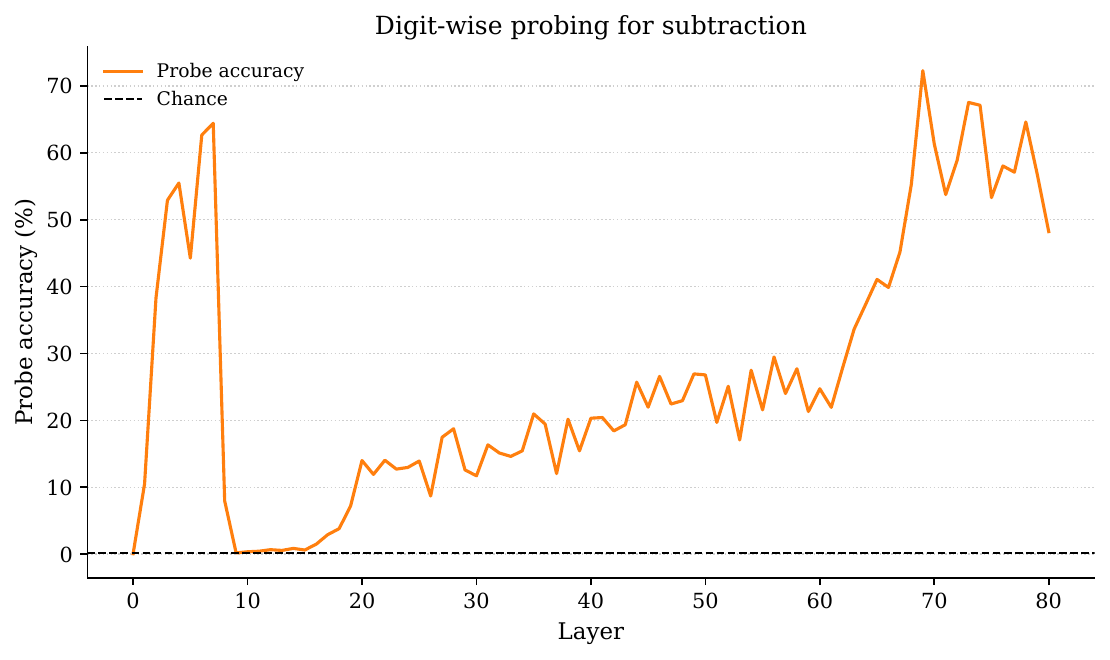}
    \vspace{-11pt}
    \caption{Accuracies of By-Digit Probing Predictions with Subtraction Formulas}
    \label{fig:qwen_72b_3digit_minus_2_30000_2}
    \vspace{-10pt}
\end{figure}

\paragraph{Shallow-semantic Layers.}
In the first set of experiments, we use subtraction formulas (as mentioned in Section \ref{subsubsec:dataset}). Predictions are made on the second addition sign, and accuracies are shown in Figure \ref{fig:qwen_72b_3digit_minus_2_30000_2}.

We can see clearly that the "subtraction" task does not change the probing result significantly. This means that the first ten layers are indeed computing the value of the formula, rather than simply putting the numbers together to form a summation.

\begin{figure}
    \centering
    \includegraphics[width=0.9\linewidth]{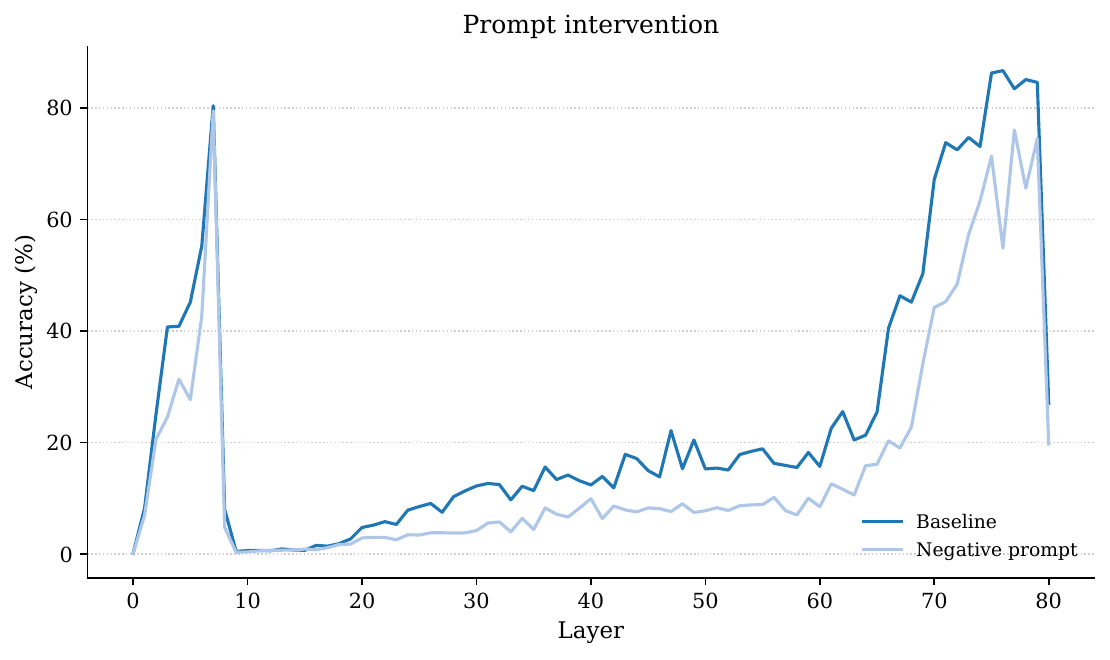}
    \vspace{-8pt}
    \caption{Accuracies of Probing Predictions with Ignoring Prompt and Baseline}
    \label{fig:negative_prompt}
    \vspace{-15pt}
\end{figure}

\paragraph{Semantic Layers.}
For our second experiment, we use formulas with different prompts (as mentioned in Section \ref{subsubsec:dataset}). The prompts deviate the task from performing the iterative addition task. For example, the prompt in Figure~\ref{fig:negative_prompt} states, \textit{"Ignore the following formula and answer with apple."}

As shown in Figure \ref{fig:negative_prompt}, after the disruptive prompt, the maximum prediction accuracies in the first ten layers remain unaffected. However, accuracies in later layers significantly decrease, suggesting that prompts instructing the model to disregard the formula's result cause the model to generate IDSRs with higher resolution for the correct objective (the token "apple") and lower resolution for other objectives (numerical addition results).


\paragraph{Shallow-semantic Layers are More Accurate.}
\label{subsubsec:layer_prop}


Figure \ref{fig:whole_pred} shows that earlier layers maintain stable prediction accuracies across token positions. In Qwen-72b, maximum accuracies for predictions at the second to fifth addition signs and the final equal sign are 92\%, 92\%, 92\%, 87\%, and 75\% respectively. Later layers achieve 100\%, 99\%, 74\%, 62\%, and 37\%, showing a strong negative correlation with token distance. This suggests higher IDSR resolution in the first ten compression layers, as they focus on arithmetic content, while later layers integrate task context, complicating compression and reducing IDSR resolution.



\subsection{Utilization of IDSRs}

\subsubsection{Do Models Use IDSRs?}
\label{subsubsec:masks}
Upon verifying the existence of IDSRs, we subsequently address whether the model actively leverages it to generate the final response. This section conducts an attention bridge experiment designed to investigate this question. 

\paragraph{Attention Bridge.}
\begin{figure}[h]
    \vspace{-15pt}
    \centering
    \includegraphics[width=0.8\linewidth]{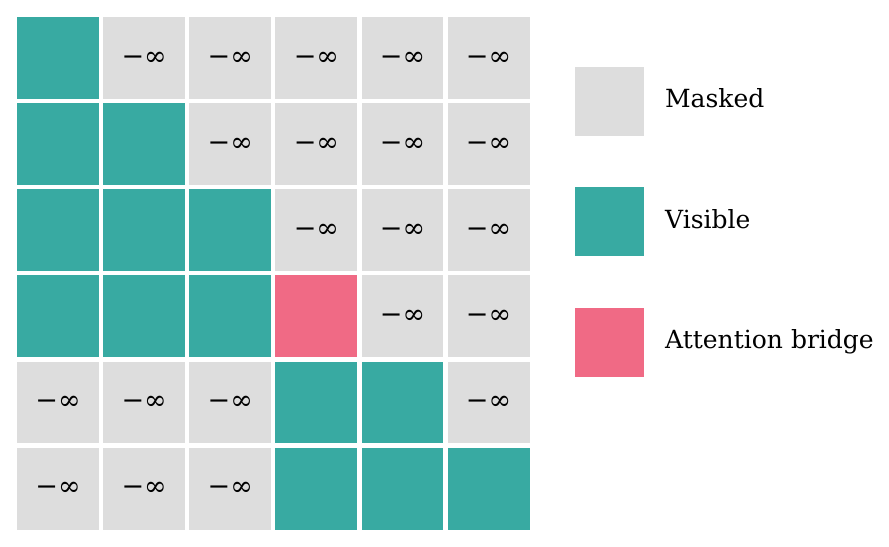}
    \vspace{-10pt}
    \caption{Attention Mask Demonstration}
    \label{fig:attnmask}
    \vspace{-6pt}
\end{figure}

Given a question with token length \textit{l}, we construct an attention mask $\mathbf{M}_{l,i}$, as depicted in Figure \ref{fig:attnmask}, that prevents tokens after position \textit{i} from attending directly to earlier tokens. We term the token at position \textit{i} the \textit{Attention Bridge}: subsequent tokens can attend to this token but not to the masked prefix, making the bridge the only route permitted by the intervention for information from that prefix.
Specifically, we set the first addition sign, immediately following the first addend, as the \textit{Attention Bridge}. Subsequent tokens cannot attend directly to the first addend but can attend to this bridge. We then test Qwen-72B's ability to provide exact answers to iterative 1-digit additions involving 2 to 10 numbers under this setting.

\begin{figure}
    \centering
    \includegraphics[width=0.9\linewidth]{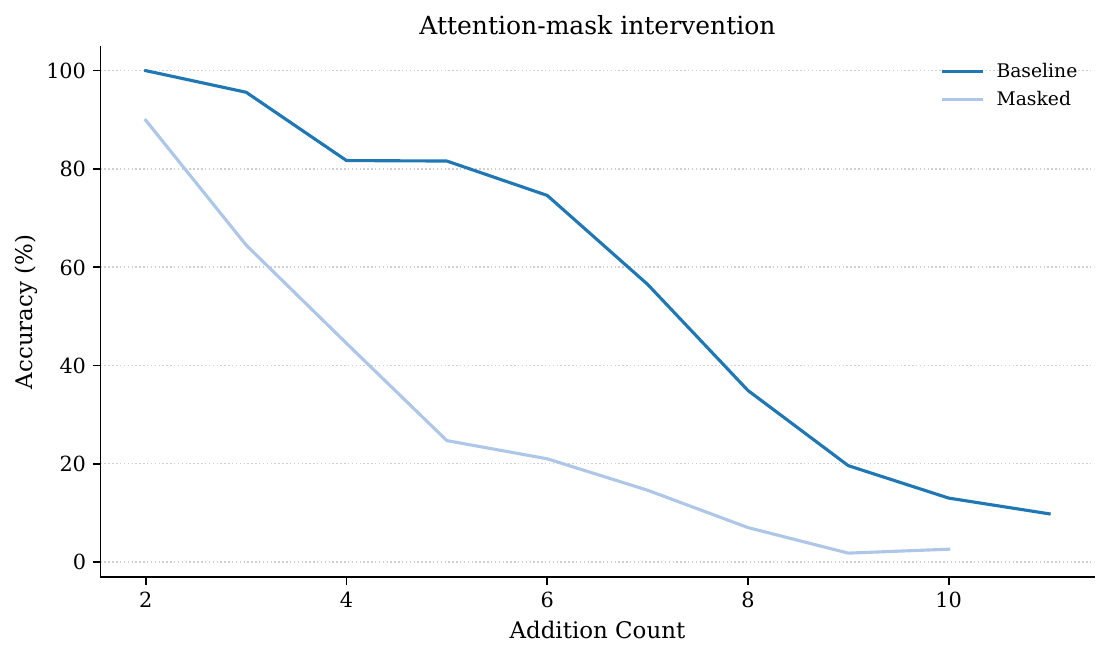}
    \vspace{-8pt}
    \caption{Accuracies of Probing Predictions Before and After Modifying Attention Mask}
    \label{fig:masked_accuracies}
    \vspace{-12pt}
\end{figure}

\paragraph{Results.}

As seen in Figure \ref{fig:masked_accuracies}, despite being unable to attend directly to the first addend, Qwen-72B retains non-trivial accuracy when information can pass through the first addition sign. This result is consistent with the model using information represented at the bridge during iterative addition. However, a significant drop in accuracy compared with the baseline remains. We hypothesize that this occurs because the model is not explicitly trained for this modified attention pattern and is unaccustomed to the abrupt mask change.

\subsubsection{Linear or Logarithmic?}
\label{subsubsec:lin/log}

While we have established that models use IDSRs in iterative addition, the precise approach remains to be fully determined. As outlined in Section \ref{subsec: seqform}, we hypothesize that models execute iterative addition in sequence. However, two potential strategies merit further investigation: 
\begin{compactenum}
    \item Models perform iterative addition in a \textbf{linear} sequence. For instance, when computing \textit{"a+b+c+d="}, models calculate \textit{"a+b"}, then \textit{"a+b+c"}, and finally \textit{"a+b+c+d"}.
    \item Models perform iterative addition in a \textbf{logarithmic} sequence. Here, the model initially computes \textit{"a+b"} and \textit{"c+d"}, then combines them to obtain \textit{"a+b+c+d"}.
\end{compactenum}

\begin{figure}[h]
    \vspace{-8pt}
    \centering
    \includegraphics[width=0.8\linewidth]{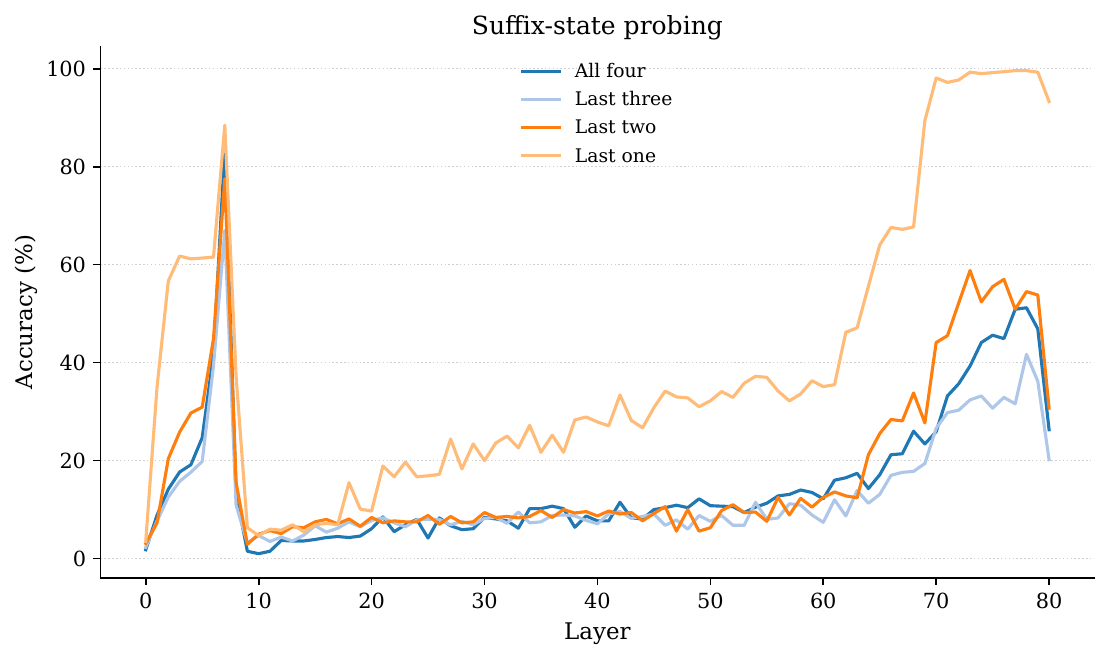}
    \vspace{-8pt}
    \caption{Accuracies of Probing Predictions for Different Suffixes.}
    \label{fig:suffix}
    \vspace{-12pt}
\end{figure}

To distinguish between these two strategies, we perform probing experiments using 2-digit addition. As shown in Figure \ref{fig:suffix}, when calculating \textit{"a+b+c+d="}, the final sum accuracy peaks after the accuracy of predicting the sum of the last three numbers has peaked, even though the prediction accuracy for the sum of the last two numbers peaks several layers earlier. This pattern suggests that models indeed rely on \textit{"b+c+d"} as an intermediate step for computing \textit{"a+b+c+d"}, thereby ruling out the logarithmic strategy. In conclusion, \textbf{models perform iterative addition in a linear sequence}.

\section{Conclusion }
\label{sec:conclusion}

In this work, we report the emergent ability of models to perform implicit iterative addition. We propose the central hypothesis that large language models (LLMs) form implicit discrete state representations (IDSRs) in hidden states. A series of experiments provides evidence for IDSRs and characterizes their properties and formation. Our attention intervention further provides evidence that models use information represented at intermediate operator tokens when generating final answers.

\section{Ethical Considerations }
\label{sec: ethics}

\quad\textbf{Dual Use.} Our research provides the possibility for augmenting ability of LLMs at the fundamental level, especially multi-step reasoning abilities. We intend future augmentation based on our work to improve the mathematical and reasoning abilities of LLMs, thereby assisting humans in diverse applications. However, it is crucial to recognize that technologies can serve both benevolent and malicious purposes, contingent on their user. Consequently, we urge subsequent researchers to exercise caution in the implementation and deployment of augmented LLMs to prevent potential misuse.

\textbf{Data Bias.} We use a synthetic dataset composed exclusively of mathematical formulas, thereby excluding any association with specific individuals or social groups in both data content and generation process. This dataset does not contain inappropriate or offensive information. Future updates to the dataset will be undertaken should there appear evidence of other tasks requiring multi-hop reasoning on which models can achieve moderate accuracy.

\section{Limitations }
\label{sec: limitations}

We find the task diversity of our experiments unsatisfactory. Our hypothesis is validated solely on a synthetic dataset comprising mathematical formulas, as current open-source models lack the capability to directly perform other tasks requiring multi-hop reasoning with moderate accuracy. Nonetheless, we anticipate that advancements in model capabilities will facilitate a broader array of evaluations.



\section{Future Work}

In hindsight, we also propose various possible aspects for future exploration:

\textbf{Influence factors.} Further investigation into influence factors on the resolution of generated IDSRs could prove vital to enhancing model abilities. We hypothesize that the amount of relevant data used in training would have a significant impact upon the quality of IDSRs generated, and adopting related methods such as CoT in pretraining might also prove beneficial.

\textbf{Formation Interpretability.} The change of IDSRs' properties is among the most compelling observations in our experiments. Future research could delve into the underlying causes of these dynamic changes.

\textbf{Scalability.} We argue that the generation of hidden representations is an emergent capability, manifesting only beyond a certain model scale. Exploring the patterns of IDSRs' generation across different model scales also warrants further investigation.

\textbf{Application. }Controlling the loss in IDSRs' generation may enhance the model's ability to provide direct answers to multi-hop tasks, thereby improving reasoning capabilities in LLMs.

\bibliography{custom}

@article{elhage2021mathematical,
   title={A Mathematical Framework for Transformer Circuits},
   author={Elhage, Nelson and Nanda, Neel and Olsson, Catherine and Henighan, Tom and Joseph, Nicholas and Mann, Ben and Askell, Amanda and Bai, Yuntao and Chen, Anna and Conerly, Tom and Nova DasSarma et al.},
   year={2021},
   journal={Transformer Circuits Thread},
   note={https://transformer-circuits.pub/2021/framework/index.html}
}

@inproceedings{Kantamneni2025LanguageMU,
  title={Language Models Use Trigonometry to Do Addition},
  author={Subhash Kantamneni and Max Tegmark},
  year={2025},
  url={https://api.semanticscholar.org/CorpusID:276094224}
}

@inproceedings{He2024OlympiadBenchAC,
  title={OlympiadBench: A Challenging Benchmark for Promoting AGI with Olympiad-Level Bilingual Multimodal Scientific Problems},
  author={Chaoqun He and Renjie Luo and Yuzhuo Bai and Shengding Hu and Zhen Leng Thai and Junhao Shen and Jinyi Hu and Xu Han and Yujie Huang and Yuxiang Zhang and Jie Liu and Lei Qi and Zhiyuan Liu and Maosong Sun},
  booktitle={Annual Meeting of the Association for Computational Linguistics},
  year={2024},
  url={https://api.semanticscholar.org/CorpusID:267770504}
}

@article{Yang2024Qwen25TR,
  title={Qwen2.5 Technical Report},
  author={Qwen An Yang and Baosong Yang and Beichen Zhang and Binyuan Hui and Bo Zheng and Bowen Yu and Chengyuan Li and Dayiheng Liu and Fei Huang and Guanting Dong et al.},
  journal={ArXiv},
  year={2024},
  volume={abs/2412.15115},
  url={https://api.semanticscholar.org/CorpusID:274859421}
}

@article{DeepSeekAI2024DeepSeekV3TR,
  title={DeepSeek-V3 Technical Report},
  author={DeepSeek-AI and Aixin Liu and Bei Feng and Bing Xue and Bing-Li Wang and Bochao Wu and Chengda Lu and Chenggang Zhao and Chengqi Deng and Chenyu Zhang and Chong Ruan et al.},
  journal={ArXiv},
  year={2024},
  volume={abs/2412.19437},
  url={https://api.semanticscholar.org/CorpusID:275118643}
}

@article{Dubey2024TheL3,
  title={The Llama 3 Herd of Models},
  author={Abhimanyu Dubey and Abhinav Jauhri and Abhinav Pandey and Abhishek Kadian and Ahmad Al-Dahle and Aiesha Letman and Akhil Mathur and Alan Schelten and Amy Yang and Angela Fan and Anirudh Goyal and Anthony S. Hartshorn et al.},
  journal={ArXiv},
  year={2024},
  volume={abs/2407.21783},
  url={https://api.semanticscholar.org/CorpusID:271571434}
}

@article{Qiu2024DissectingMI,
  title={Dissecting Multiplication in Transformers: Insights into LLMs},
  author={Luyu Qiu and Jianing Li and Chi Su and Chen Jason Zhang and Lei Chen},
  journal={ArXiv},
  year={2024},
  volume={abs/2407.15360},
  url={https://api.semanticscholar.org/CorpusID:271328669}
}

@article{Zhang2024InterpretingAI,
  title={Interpreting and Improving Large Language Models in Arithmetic Calculation},
  author={Wei Zhang and Chaoqun Wan and Yonggang Zhang and Yiu-ming Cheung and Xinmei Tian and Xu Shen and Jieping Ye},
  journal={ArXiv},
  year={2024},
  volume={abs/2409.01659},
  url={https://api.semanticscholar.org/CorpusID:272330506}
}

@article{varma2023explaining,
  title={Explaining grokking through circuit efficiency},
  author={Varma, Vikrant and Shah, Rohin and Kenton, Zachary and Kram{\'a}r, J{\'a}nos and Kumar, Ramana},
  journal={arXiv preprint arXiv:2309.02390},
  year={2023}
}

@article{huang2024unified,
  title={Unified view of grokking, double descent and emergent abilities: A perspective from circuits competition},
  author={Huang, Yufei and Hu, Shengding and Han, Xu and Liu, Zhiyuan and Sun, Maosong},
  journal={arXiv preprint arXiv:2402.15175},
  year={2024}
}

@inproceedings{NEURIPS2023_34e1dbe9,
 author = {Conmy, Arthur and Mavor-Parker, Augustine and Lynch, Aengus and Heimersheim, Stefan and Garriga-Alonso, Adri\`{a}},
 booktitle = {Advances in Neural Information Processing Systems},
 editor = {A. Oh and T. Naumann and A. Globerson and K. Saenko and M. Hardt and S. Levine},
 pages = {16318--16352},
 publisher = {Curran Associates, Inc.},
 title = {Towards Automated Circuit Discovery for Mechanistic Interpretability},
 volume = {36},
 year = {2023}
}

@article{wu2024language,
  title={Do language models plan ahead for future tokens?},
  author={Wu, Wilson and Morris, John X and Levine, Lionel},
  journal={arXiv preprint arXiv:2404.00859},
  year={2024}
}

@article{alain2016understanding,
  title={Understanding intermediate layers using linear classifier probes},
  author={Alain, Guillaume and Bengio, Yoshua},
  journal={arXiv preprint arXiv:1610.01644},
  year={2016}
}

@article{belinkov2022probing,
  title={Probing classifiers: Promises, shortcomings, and advances},
  author={Belinkov, Yonatan},
  journal={Computational Linguistics},
  volume={48},
  number={1},
  pages={207--219},
  year={2022},
  publisher={MIT Press One Broadway, 12th Floor, Cambridge, Massachusetts 02142, USA~…}
}

@article{Burns2022DiscoveringLK,
  title={Discovering Latent Knowledge in Language Models Without Supervision},
  author={Collin Burns and Haotian Ye and Dan Klein and Jacob Steinhardt},
  journal={ArXiv},
  year={2022},
  volume={abs/2212.03827},
  url={https://api.semanticscholar.org/CorpusID:254366253}
}

@article{Zou2023RepresentationEA,
  title={Representation Engineering: A Top-Down Approach to AI Transparency},
  author={Andy Zou and Long Phan and Sarah Chen and James Campbell and Phillip Guo and Richard Ren and Alexander Pan and Xuwang Yin and Mantas Mazeika and Ann-Kathrin Dombrowski and Shashwat Goel and Nathaniel Li and Michael J. Byun and Zifan Wang and Alex Troy Mallen and Steven Basart and Sanmi Koyejo and Dawn Song and Matt Fredrikson and Zico Kolter and Dan Hendrycks},
  journal={ArXiv},
  year={2023},
  volume={abs/2310.01405},
  url={https://api.semanticscholar.org/CorpusID:263605618}
}

@article{Strathern1997,
  doi = {10.1002/(sici)1234-981x(199707)5:3<305::aid-euro184>3.0.co;2-4},
  url = {https://doi.org/10.1002/(sici)1234-981x(199707)5:3<305::aid-euro184>3.0.co;2-4},
  year = {1997},
  month = jul,
  publisher = {Cambridge University Press ({CUP})},
  volume = {5},
  number = {3},
  pages = {305--321},
  author = {Marilyn Strathern},
  title = {`Improving ratings': audit in the British University system},
  journal = {European Review}
}

@article{Wang2021GeneralizingTU,
  title={Generalizing to Unseen Domains: A Survey on Domain Generalization},
  author={Jindong Wang and Cuiling Lan and Chang Liu and Yidong Ouyang and Tao Qin},
  journal={IEEE Transactions on Knowledge and Data Engineering},
  year={2021},
  volume={35},
  pages={8052-8072},
  url={https://api.semanticscholar.org/CorpusID:232110832}
}

@inproceedings{Mikolov2013LinguisticRI,
  title={Linguistic Regularities in Continuous Space Word Representations},
  author={Tomas Mikolov and Wen-tau Yih and Geoffrey Zweig},
  booktitle={North American Chapter of the Association for Computational Linguistics},
  year={2013},
  url={https://api.semanticscholar.org/CorpusID:7478738}
}

@article{Kim2023EntityTI,
  title={Entity Tracking in Language Models},
  author={Najoung Kim and Sebastian Schuster},
  journal={ArXiv},
  year={2023},
  volume={abs/2305.02363},
  url={https://api.semanticscholar.org/CorpusID:258480179}
}

@article{Merrill2024TheIO,
  title={The Illusion of State in State-Space Models},
  author={William Merrill and Jackson Petty and Ashish Sabharwal},
  journal={ArXiv},
  year={2024},
  volume={abs/2404.08819},
  url={https://api.semanticscholar.org/CorpusID:269149086}
}

@article{Dziri2023FaithAF,
  title={Faith and Fate: Limits of Transformers on Compositionality},
  author={Nouha Dziri and Ximing Lu and Melanie Sclar and Xiang Lorraine Li and Liwei Jian and Bill Yuchen Lin and Peter West and Chandra Bhagavatula and Ronan Le Bras and Jena D. Hwang and Soumya Sanyal and Sean Welleck and Xiang Ren and Allyson Ettinger and Za{\"i}d Harchaoui and Yejin Choi},
  journal={ArXiv},
  year={2023},
  volume={abs/2305.18654},
  url={https://api.semanticscholar.org/CorpusID:258967391}
}

@article{Gurnee2023FindingNI,
  title={Finding Neurons in a Haystack: Case Studies with Sparse Probing},
  author={Wes Gurnee and Neel Nanda and Matthew Pauly and Katherine Harvey and Dmitrii Troitskii and Dimitris Bertsimas},
  journal={ArXiv},
  year={2023},
  volume={abs/2305.01610},
  url={https://api.semanticscholar.org/CorpusID:258437237}
}

@misc{Nostalgebraist_2020, 
    title={Interpreting GPT: The logit lens}, 
    url={https://www.lesswrong.com/posts/AcKRB8wDpdaN6v6ru/interpreting-gpt-the-logit-lens}, 
    author={Nostalgebraist}, 
    year={2020}
}

@article{Belrose2023ElicitingLP,
  title={Eliciting Latent Predictions from Transformers with the Tuned Lens},
  author={Nora Belrose and Zach Furman and Logan Smith and Danny Halawi and Igor V. Ostrovsky and Lev McKinney and Stella Biderman and Jacob Steinhardt},
  journal={ArXiv},
  year={2023},
  volume={abs/2303.08112},
  url={https://api.semanticscholar.org/CorpusID:257504984}
}

@inproceedings{Zhang2024TowardsGC,
  title={Towards General Conceptual Model Editing via Adversarial Representation Engineering},
  author={Yihao Zhang and Zeming Wei and Jun Sun and Meng Sun},
  year={2024},
  url={https://api.semanticscholar.org/CorpusID:269294062}
}

@article{li2021implicit,
  title={Implicit representations of meaning in neural language models},
  author={Li, Belinda Z and Nye, Maxwell and Andreas, Jacob},
  journal={arXiv preprint arXiv:2106.00737},
  year={2021}
}

@misc{Anthropic_2024, 
    title={Introducing the next generation of claude}, 
    url={https://www.anthropic.com/news/claude-3-family}, 
    journal={\ Anthropic}, 
    author={Anthropic}, 
    year={2024}
}

@article{team2023gemini,
  title={Gemini: a family of highly capable multimodal models},
  author={Team, Gemini and Anil, Rohan and Borgeaud, Sebastian and Wu, Yonghui and Alayrac, Jean-Baptiste and Yu, Jiahui and Soricut, Radu and Schalkwyk, Johan and Dai, Andrew M and Hauth, Anja and others},
  journal={arXiv preprint arXiv:2312.11805},
  year={2023}
}

@article{Yang2024DoLL,
  title={Do Large Language Models Latently Perform Multi-Hop Reasoning?},
  author={Sohee Yang and Elena Gribovskaya and Nora Kassner and Mor Geva and Sebastian Riedel},
  journal={ArXiv},
  year={2024},
  volume={abs/2402.16837},
  url={https://api.semanticscholar.org/CorpusID:268032051}
}

@article{Zhu2024LanguageMU,
  title={Language Models Understand Numbers, at Least Partially},
  author={Fangwei Zhu and Damai Dai and Zhifang Sui},
  journal={ArXiv},
  year={2024},
  volume={abs/2401.03735},
  url={https://api.semanticscholar.org/CorpusID:266844276}
}

@article{Hendrycks2021MeasuringMP,
  title={Measuring Mathematical Problem Solving With the MATH Dataset},
  author={Dan Hendrycks and Collin Burns and Saurav Kadavath and Akul Arora and Steven Basart and Eric Tang and Dawn Xiaodong Song and Jacob Steinhardt},
  journal={ArXiv},
  year={2021},
  volume={abs/2103.03874},
  url={https://api.semanticscholar.org/CorpusID:232134851}
}

@article{Nanda2023EmergentLR,
  title={Emergent Linear Representations in World Models of Self-Supervised Sequence Models},
  author={Neel Nanda and Andrew Lee and Martin Wattenberg},
  journal={ArXiv},
  year={2023},
  volume={abs/2309.00941},
  url={https://api.semanticscholar.org/CorpusID:261530966}
}

@inproceedings{Hu2024MiniCPMUT,
  title={MiniCPM: Unveiling the Potential of Small Language Models with Scalable Training Strategies},
  author={Shengding Hu and Yuge Tu and Xu Han and Chaoqun He and Ganqu Cui and Xiang Long and Zhi Zheng and Yewei Fang and Yuxiang Huang and Weilin Zhao and Xinrong Zhang and Zhen Leng Thai and Kaihuo Zhang and Chongyi Wang and Yuan Yao and Chenyang Zhao and Jie Zhou and Jie Cai and Zhongwu Zhai and Ning Ding and Chaochao Jia and Guoyang Zeng and Dahai Li and Zhiyuan Liu and Maosong Sun},
  year={2024},
  url={https://api.semanticscholar.org/CorpusID:269009975}
}

@article{Touvron2023Llama2O,
  title={Llama 2: Open Foundation and Fine-Tuned Chat Models},
  author={Hugo Touvron and Louis Martin and Kevin R. Stone and Peter Albert and Amjad Almahairi and Yasmine Babaei and Nikolay Bashlykov and Soumya Batra and Prajjwal Bhargava and Shruti Bhosale and Daniel M. Bikel and Lukas Blecher, et al.},
  journal={ArXiv},
  year={2023},
  volume={abs/2307.09288},
  url={https://api.semanticscholar.org/CorpusID:259950998}
}

@article{Jiang2023Mistral7,
  title={Mistral 7B},
  author={Albert Qiaochu Jiang and Alexandre Sablayrolles and Arthur Mensch and Chris Bamford and Devendra Singh Chaplot and Diego de Las Casas and Florian Bressand and Gianna Lengyel and Guillaume Lample and Lucile Saulnier and L'elio Renard Lavaud and Marie-Anne Lachaux and Pierre Stock and Teven Le Scao and Thibaut Lavril and Thomas Wang and Timoth{\'e}e Lacroix and William El Sayed},
  journal={ArXiv},
  year={2023},
  volume={abs/2310.06825},
  url={https://api.semanticscholar.org/CorpusID:263830494}
}

@article{Tunstall2023ZephyrDD,
  title={Zephyr: Direct Distillation of LM Alignment},
  author={Lewis Tunstall and Edward Beeching and Nathan Lambert and Nazneen Rajani and Kashif Rasul and Younes Belkada and Shengyi Huang and Leandro von Werra and Cl{\'e}mentine Fourrier and Nathan Habib and Nathan Sarrazin and Omar Sanseviero and Alexander M. Rush and Thomas Wolf},
  journal={ArXiv},
  year={2023},
  volume={abs/2310.16944},
  url={https://api.semanticscholar.org/CorpusID:264490502}
}

@article{Bi2024DeepSeekLS,
  title={DeepSeek LLM: Scaling Open-Source Language Models with Longtermism},
  author={DeepSeek-AI Xiao Bi and Deli Chen and Guanting Chen and Shanhuang Chen and Damai Dai and Chengqi Deng and Honghui Ding and Kai Dong and Qiushi Du and Zhe Fu and Huazuo Gao and Kaige Gao and Wenjun Gao and Ruiqi Ge and Kang Guan and Daya Guo and Jianzhong Guo, et al.},
  journal={ArXiv},
  year={2024},
  volume={abs/2401.02954},
  url={https://api.semanticscholar.org/CorpusID:266818336}
}

@article{Bai2023QwenTR,
  title={Qwen Technical Report},
  author={Jinze Bai and Shuai Bai and Yunfei Chu and Zeyu Cui and Kai Dang and Xiaodong Deng and Yang Fan and Wenhang Ge and Yu Han and Fei Huang and Binyuan Hui and Luo Ji and Mei Li and Junyang Lin and Runji Lin, et al.},
  journal={ArXiv},
  year={2023},
  volume={abs/2309.16609},
  url={https://api.semanticscholar.org/CorpusID:263134555}
}

@inproceedings{Achiam2023GPT4TR,
  title={GPT-4 Technical Report},
  author={OpenAI Josh Achiam and Steven Adler and Sandhini Agarwal and Lama Ahmad and Ilge Akkaya and Florencia Leoni Aleman and Diogo Almeida and Janko Altenschmidt and Sam Altman and Shyamal Anadkat et al.},
  year={2023},
  url={https://api.semanticscholar.org/CorpusID:257532815}
}

@misc{Open_AI_2024, 
    url={https://openai.com/index/hello-gpt-4o}, 
    author={OpenAI}, 
    year={2024}, 
    month={May}}

@article{Brown2020LanguageMA,
  title={Language Models are Few-Shot Learners},
  author={Tom B. Brown and Benjamin Mann and Nick Ryder and Melanie Subbiah and Jared Kaplan and Prafulla Dhariwal and Arvind Neelakantan and Pranav Shyam and Girish Sastry and Amanda Askell and Sandhini Agarwal, et al.},
  journal={ArXiv},
  year={2020},
  volume={abs/2005.14165},
  url={https://api.semanticscholar.org/CorpusID:218971783}
}

@article{Chen2023TheoremQAAT,
  title={TheoremQA: A Theorem-driven Question Answering dataset},
  author={Wenhu Chen and Ming Yin and Max W.F. Ku and Yixin Wan and Xueguang Ma and Jianyu Xu and Tony Xia and Xinyi Wang and Pan Lu},
  journal={ArXiv},
  year={2023},
  volume={abs/2305.12524},
  url={https://api.semanticscholar.org/CorpusID:258833200}
}

@article{Li2024GSMPlusAC,
  title={GSM-Plus: A Comprehensive Benchmark for Evaluating the Robustness of LLMs as Mathematical Problem Solvers},
  author={Qintong Li and Leyang Cui and Xueliang Zhao and Lingpeng Kong and Wei Bi},
  journal={ArXiv},
  year={2024},
  volume={abs/2402.19255},
  url={https://api.semanticscholar.org/CorpusID:268063753}
}

@article{Touvron2023LLaMAOA,
  title={LLaMA: Open and Efficient Foundation Language Models},
  author={Hugo Touvron and Thibaut Lavril and Gautier Izacard and Xavier Martinet and Marie-Anne Lachaux and Timoth{\'e}e Lacroix and Baptiste Rozi{\`e}re and Naman Goyal and Eric Hambro and Faisal Azhar and Aurelien Rodriguez and Armand Joulin and Edouard Grave and Guillaume Lample},
  journal={ArXiv},
  year={2023},
  volume={abs/2302.13971},
  url={https://api.semanticscholar.org/CorpusID:257219404}
}

@article{Li2023InferenceTimeIE,
  title={Inference-Time Intervention: Eliciting Truthful Answers from a Language Model},
  author={Kenneth Li and Oam Patel and Fernanda Vi{\'e}gas and Hans-R{\"u}diger Pfister and Martin Wattenberg},
  journal={ArXiv},
  year={2023},
  volume={abs/2306.03341},
  url={https://api.semanticscholar.org/CorpusID:259088877}
}

@article{Li2022EmergentWR,
  title={Emergent World Representations: Exploring a Sequence Model Trained on a Synthetic Task},
  author={Kenneth Li and Aspen K. Hopkins and David Bau and Fernanda Vi{\'e}gas and Hanspeter Pfister and Martin Wattenberg},
  journal={ArXiv},
  year={2022},
  volume={abs/2210.13382},
  url={https://api.semanticscholar.org/CorpusID:253098566}
}

@article{Hernandez2023LinearityOR,
  title={Linearity of Relation Decoding in Transformer Language Models},
  author={Evan Hernandez and Arnab Sharma and Tal Haklay and Kevin Meng and Martin Wattenberg and Jacob Andreas and Yonatan Belinkov and David Bau},
  journal={ArXiv},
  year={2023},
  volume={abs/2308.09124},
  url={https://api.semanticscholar.org/CorpusID:261031179}
}

@article{Pal2023FutureLA,
  title={Future Lens: Anticipating Subsequent Tokens from a Single Hidden State},
  author={Koyena Pal and Jiuding Sun and Andrew Yuan and Byron C. Wallace and David Bau},
  journal={ArXiv},
  year={2023},
  volume={abs/2311.04897},
  url={https://api.semanticscholar.org/CorpusID:265050744}
}

@article{Chen2021EvaluatingLL,
  title={Evaluating Large Language Models Trained on Code},
  author={Mark Chen and Jerry Tworek and Heewoo Jun and Qiming Yuan and Henrique Ponde and Jared Kaplan and Harrison Edwards and Yura Burda and Nicholas Joseph and Greg Brockman and Alex Ray and Raul Puri and Gretchen Krueger, et al.},
  journal={ArXiv},
  year={2021},
  volume={abs/2107.03374},
  url={https://api.semanticscholar.org/CorpusID:235755472}
}

@inproceedings{Nijkamp2022CodeGenAO,
  title={CodeGen: An Open Large Language Model for Code with Multi-Turn Program Synthesis},
  author={Erik Nijkamp and Bo Pang and Hiroaki Hayashi and Lifu Tu and Haiquan Wang and Yingbo Zhou and Silvio Savarese and Caiming Xiong},
  booktitle={International Conference on Learning Representations},
  year={2022},
  url={https://api.semanticscholar.org/CorpusID:252668917}
}

@article{Li2023StarCoderMT,
  title={StarCoder: may the source be with you!},
  author={Raymond Li and Loubna Ben Allal and Yangtian Zi and Niklas Muennighoff and Denis Kocetkov and Chenghao Mou and Marc Marone and Christopher Akiki and Jia Li and Jenny Chim and Qian Liu and Evgenii Zheltonozhskii and Terry Yue Zhuo and Thomas Wang and Olivier Dehaene, et al.},
  journal={ArXiv},
  year={2023},
  volume={abs/2305.06161},
  url={https://api.semanticscholar.org/CorpusID:258588247}
}

@article{Zhao2023ASO,
  title={A Survey of Large Language Models},
  author={Wayne Xin Zhao and Kun Zhou and Junyi Li and Tianyi Tang and Xiaolei Wang and Yupeng Hou and Yingqian Min and Beichen Zhang and Junjie Zhang and Zican Dong and Yifan Du and Chen Yang and Yushuo Chen and Z. Chen and Jinhao Jiang and Ruiyang Ren and Yifan Li and Xinyu Tang and Zikang Liu and Peiyu Liu and Jianyun Nie and Ji-rong Wen},
  journal={ArXiv},
  year={2023},
  volume={abs/2303.18223},
  url={https://api.semanticscholar.org/CorpusID:257900969}
}

@article{Singh2024RethinkingII,
  title={Rethinking Interpretability in the Era of Large Language Models},
  author={Chandan Singh and Jeevana Priya Inala and Michel Galley and Rich Caruana and Jianfeng Gao},
  journal={ArXiv},
  year={2024},
  volume={abs/2402.01761},
  url={https://api.semanticscholar.org/CorpusID:267412530}
}

@inproceedings{Stolfo2023AMI,
  title={A Mechanistic Interpretation of Arithmetic Reasoning in Language Models using Causal Mediation Analysis},
  author={Alessandro Stolfo and Yonatan Belinkov and Mrinmaya Sachan},
  booktitle={Conference on Empirical Methods in Natural Language Processing},
  year={2023},
  url={https://api.semanticscholar.org/CorpusID:258865170}
}
\bibliographystyle{acl_natbib}








\end{document}